\documentclass[12pt,onecolumn]{article}	% originally amsart, the default for mathematical articles.
\usepackage[top=1in, bottom=1in, left=1in, right=1in]{geometry}
\usepackage{graphicx}
\usepackage{amssymb}
\usepackage{amsmath}  				% I added this per tutorial recommendation
\usepackage{epstopdf}  				% works with graphicx to convert epst into pdf
\usepackage{graphicx,float,wrapfig} 	% for embedding graphics
\usepackage{color}					% Allows picking of custom RGB colors
\usepackage[colorlinks,linktocpage=true]{hyperref}		% for hotlinking all \cite{} and \ref{}
\usepackage[font=small,labelfont=bf,textfont=sf,
labelsep=quad]{caption}				% points hyperref to top of figure, not caption
\usepackage{tikz}
\definecolor{purple}{RGB}{152, 66, 227}
\definecolor{magenta}{RGB}{205, 16,118}
\definecolor{turquoise}{RGB}{53, 173, 153}
\definecolor{gray}{RGB}{200, 200, 200}

\makeatletter
\newcommand{\getcitenumber}[1]{% \getcitenumber{<cite>}
  \@ifundefined{b@#1}% Check if <cite> exists
    {\hbox{\reset@font\bfseries ?}}% <cite> doesn't exist; print ?
    {\csname b@#1\endcsname}}% <cite> exists; print `\b@<cite>
\makeatother

\usepackage[superscript,space]{cite}	

\hypersetup{						% Hyperlinked reference setup; can use RGB color values
  citecolor= black,
  linkcolor=black,
}
\usepackage{setspace}   

\usepackage [english]{babel}
\usepackage [english = american]{csquotes}
\MakeOuterQuote{"}

\onehalfspace                   				%   change line spacing

 \usepackage[protrusion=false,expansion=false]{microtype}

\renewcommand{\v}[1]{\ensuremath{\mathbf{#1}}} % for vectors
\newcommand{\gv}[1]{\ensuremath{\mbox{\boldmath$ #1 $}}} 
\newcommand{\abs}[1]{\left| #1 \right|} % for absolute value
\let\baraccent=\= % rename builtin command \= to \baraccent
\renewcommand{\=}[1]{\stackrel{#1}{=}} % for putting numbers above =
\renewcommand\eqref[1]{Eq.\;\ref{#1}} % new version of eqref

\begin{document}

%\title{Generative machine learning and chaotic dynamics}
%\title{The nonlinear dynamics of generative machine learning}
%\title{Generative learning models and chaotic dynamics}
%\title{The nonlinear dynamics of generative learning models}
\title{Generative learning for nonlinear dynamics}

\author
{William Gilpin$^{1,2\ast}$\\
\\
\normalsize{$^{1}$Department of Physics, $^{2}$The Oden Institute for Computational Engineering \& Sciences}\\
\normalsize{The University of Texas at Austin, Austin, TX}\\
\\
\normalsize{$^{\ast}$wgilpin@utexas.edu}
\\
}

%\noindent\today

 \date{}  % Activate to display a given date or no date

\maketitle

\newpage
%% The unstructured abstract of up to 200 words should introduce the main themes of the article in a succinct, easily digestible way to engage a broad readership. All information mentioned in the abstract must be addressed in the main article. The abstract should con- tain minimal specialist details, no references and no display item citations.
\begin{abstract}

Modern generative machine learning models demonstrate surprising ability to create realistic outputs far beyond their training data, such as photorealistic artwork, accurate protein structures, or conversational text. These successes suggest that generative models learn to effectively parametrize and sample arbitrarily complex distributions. Beginning half a century ago, foundational works in nonlinear dynamics used tools from information theory to infer properties of chaotic attractors from time series, motivating the development of algorithms for parametrizing chaos in real datasets. In this perspective, we aim to connect these classical works to emerging themes in large-scale generative statistical learning. We first consider classical attractor reconstruction, which mirrors constraints on latent representations learned by state space models of time series. We next revisit early efforts to use symbolic approximations to compare minimal discrete generators underlying complex processes, a problem relevant to modern efforts to distill and interpret black-box statistical models. Emerging interdisciplinary works bridge nonlinear dynamics and learning theory, such as operator-theoretic methods for complex fluid flows, or detection of broken detailed balance in biological datasets. We anticipate that future machine learning techniques may revisit other classical concepts from nonlinear dynamics, such as transinformation decay and complexity-entropy tradeoffs.
%Modern generative machine learning models demonstrate surprising ability to create realistic outputs far beyond their training data, suggesting that they learn effective parametrize and sample arbitrarily complex data distributions. 
%
%We anticipate that other traditional properties of chaotic systems like transinformation decay and complexity-entropy tradeoffs may inform future developments in data-driven modelling of complex systems.
\end{abstract}

\newpage

\tableofcontents
%\listoffigures
\newpage

% 3 -5 figures, minimum of 2
% Target ~15 pages including figures
\section{Introduction: Chaos as a generative process}

%Two trajectories originating from distinct locations on a strange attractor will never recur nor intersect, and so observing a chaotic system continuously produces new information about the complex geometry of its underlying attractor \cite{crutchfield1982symbolic,cvitanovic2005chaos}. This accumulated information reveals detail at ever-decreasing scales, giving rise to the fractal structures and continuous spectra that characterize chaos \cite{farmer1982information}. 
The strangeness of a strange attractor stems from its unexpected geometry, which can only be visualized by observing a chaotic system evolve for an extended duration. Chaotic systems therefore continuously produce information, which gradually reveals their characteristic fractal structure at ever-decreasing scales \cite{crutchfield1982symbolic,cvitanovic2005chaos,farmer1982information}. The notion of information production by chaotic systems inspired early efforts to frame computation as a physical theory---from Feynman's estimation of the information stored in an ideal gas \cite{feynman2018feynman}, to John Archibald Wheeler's analogies between traveling salesman algorithms and molecular chaos \cite{wheeler1983recognizing}. Wheeler would later declare "it from bit"---that physical theories ultimately encode computational primitives \cite{wheeler1992recent}.

Contemporaneous to Wheeler's remark, work by the dynamical systems community formalized information production by chaotic systems \cite{shaw1981strange,pompe1986state,crutchfield1989inferring,grassberger1991information,sauer1991embedology}. Continuous-time chaotic systems encountered in the natural world, ranging from turbulent cascades to intertwined stellar orbits, act as analogue computers that manipulate and transform information encoded in their initial conditions and parameters \cite{farmer1982information,pesin1977characteristic,gilpin2018cryptographic}. Given a chaotic dynamical system $\dot{\v{x}} = \v{f}(\v{x})$, Pesin's formula states that its entropy production rate is proportional to the sum of its positive Lyapunov exponents \cite{pesin1977characteristic}, which measure the rate at which nearby trajectories diverge along different directions on a chaotic attractor,
\[
	\mathcal{H} = \sum_{\lambda_i > 0} \lambda_i.
\]
The entropy  $\mathcal{H}$ represents the Kolmogorov-Sinai entropy, which can be estimated by coarse-graining the system's phase space with infinitesimal bins, and then calculating the probability of the system occupying each bin over an extended period \cite{sinai1972gibbs}. Pesin's formula thus relates properties of the dynamics to the rate of information production of the system $\v{f}(\v{x})$; systems with greater chaoticity more quickly reveal the points on their attractor. The formula therefore connects dynamics, attractor geometry, and information in the evolution of chaotic systems.

%\subsection{Generative models and diffusion}
Recent developments in generative statistical learning motivate revisiting information production by chaotic systems. Many machine learning algorithms implicitly estimate the underlying distribution $p(\v{x})$ based on a finite number of inputs $\v{x}$ seen during training \cite{blei2017variational}. Supervised learning algorithms seek to construct the conditional probability distribution $p(\v{y}|\v{x})$ of a target state $\v{y}$ given knowledge of an input $\v{x}$. In image classification, $\v{y}$ comprises a discrete label for an image $\v{x}$; while in forecasting, $\v{y}$ represents the future system state conditioned on the past observations $\v{x}$. In contrast, unsupervised learning constructs map between the estimated $p(\v{x})$ and a latent space $p(\v{z} | \v{x})$ in which underlying patterns in the data become apparent. Recently-popularized \textit{generative models}, like generative adversarial networks or variational autoencoders, seek to directly sample $p(\v{x})$ in order to produce new examples $\v{x}'$ resembling training data cases. These methods either directly sample a smooth estimate of $p(\v{x})$ directly constructed from the training data, or instead sample the latent space $p(\v{z}')$ and then decode the result through the inverse transform $p(\v{x}' | \v{z}')$ \cite{blei2017variational}. 

However, because generative models are often used in applications where $\v{x}$ is high-dimensional, drawing representative samples from $p(\v{x})$ usually proves difficult due to the curse of dimensionality, leading to a high sample rejection rate. Methods suitable for high-dimensional distributions such as Markov Chain Monte Carlo select the next sample $\v{x}_{i+1}$ based on the previous accepted sample $\v{x}_i$. A simplified such scheme centers a Gaussian proposal distribution around the current sample, $\v{x}_{i+1} \sim \mathcal{N}(\v{x}_i, \gv\Sigma)$, with the covariance matrix $\gv\Sigma$ aligned with the distribution's local geometry as estimated from gradient information or prior samples. A sample drawn from this proposal distribution provides an amount of local information given by
\begin{equation}
	\mathcal{H}=  \ln (2\pi \mathrm {e})^{N/2} + \sum_{i=1}^N \ln{\sigma_i}
\label{entgauss}
\end{equation}
where $\sigma_i$ denote the standard deviations of the $N$-dimensional distribution along its principal axes. Like Pesin's formula, this expression relates novelty in the form of information gain to local geometric properties of the underlying manifold. Just as chaotic systems diverge along unstable manifolds associated with positive Lyapunov exponents, complex data distributions contain flatter local directions that dominate their apparent diversity \cite{goodfellow2016deep,edelman1998geometry}.  

Consistent with this connection between dynamics and sampling, many modern large-scale generative models implement sampling schemes that simulate dynamical systems mapping input data $\v{x}$ to the latent space $\v{z}$ \cite{sohl2015deep,song2019generative}. The ability of generative models to capture probabilistic structure in complex datasets has enabled their recognizable applications in human-like text and artwork generation, and has enabled scientific applications ranging from latent variable models of spiking neurons \cite{pandarinath2018inferring,koppe2019identifying}, to upsampling turbulent flow simulations \cite{yousif2021high}. 

The recent emergence of generative models of complex datasets therefore motivates revisiting classical works on information processing in chaotic systems. Chaos and statistics have well-established connections through ergodic theory \cite{bowen1973symbolic,sinai1972gibbs,cvitanovic2005chaos}, which stimulated the development of early statistical methods for identifying the dynamical systems that act as generators for experimental time series \cite{gershenfeld1988experimentalist,grassberger1991information,abarbanel1993analysis}. Several concepts that persist today in machine learning research contain roots in these influential early works. While several recent reviews highlight work at the intersection of data-driven modeling and dynamical systems \cite{bahri2020statistical,karniadakis2021physics,brunton2022modern,mezic2013analysis,otto2021koopman,ghadami2022data}, this Perspective focuses specifically on two classical problems---representing systems given partial measurements, and inferring minimal dynamical generators of complex processes---that complement emerging work in statistical learning. We thus seek to bridge classical and emerging ideas and revisit "it from bit," by situating recent discoveries within dynamical systems theory.

\begin{figure}
{
\centering
\includegraphics[width=0.7\linewidth]{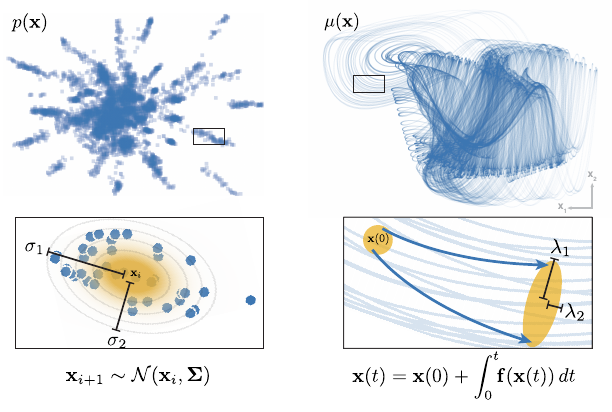}
\caption{
(Left) An example of a complex probability distribution $p(\v{x})$ and a schematic of a simplified Markov Chain Monte Carlo sampling scheme. (Right) The natural measure $\mu(\v{x})$ of a strange attractor, and a schematic of the divergence of a set of initial conditions. In these examples, $p(\v{x})$ is taken from the distribution of proteins learned by a variational autoencoder trained on amino acid sequences \cite{ding2019deciphering}, while $\mu(\v{x})$ comes from a reduced order deterministic model of a convective cell.
}
\label{generative}
}
\end{figure}

\section{Representing and propagating chaotic dynamics}

Large statistical learning models often parametrize complex datasets in low-dimensional latent spaces. For example, artificial image generators aim to invertibly map the space of natural images to a latent probability distribution \cite{song2019generative}. The empirical success of such approaches is termed the "manifold hypothesis;" that high-dimensional datasets typically cluster near low-dimensional manifolds \cite{fefferman2016testing,boumal2023introduction}. In the context of time series, successful data-driven re-parametrization implies that the dynamics arise from a low-dimensional attractor embedded within the higher-dimensional ambient space of the data. In this case, the apparent complexity of a measured time series is a matter of representation---complexity can be "transformed out" by identifying and parameterizing this structure.

\begin{figure}
{
\centering
\includegraphics[width=0.7\linewidth]{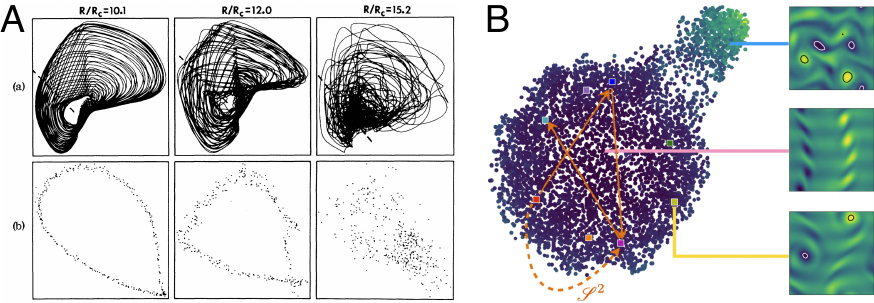}
\caption{
{\bf Latent dynamics revisit classical attractor reconstruction.} (A) Time-delay embeddings of a univariate time series representing the radial velocity of a flow, at three different Reynolds numbers ($R$) leading to turbulence. Poincare sections are shown below each embedding. (B) The latent space of an autoencoder neural network trained on weak turbulence ($R = 40$). The latent states are further embedded in two dimensions using t-distributed stochastic neighbor embedding (t-SNE). Shading indicates power dissipation, and connected states indicate equivalent flow configurations under a discrete symmetry operation.
\textit{Panel A is reprinted from Brandst{\"a}ter et al. 1983. Panel B is reprinted from Page, Brenner, and Kerswell, 2021}.\cite{brandstater1983low,page2021revealing}
}
\label{statespace}
}
\end{figure}

Historical work by the dynamical systems community sought to reconstruct the manifolds underlying dynamical systems based on limited observations. Contemporaneously to the introduction of Pesin's formula, several works formulated methods for reconstructing dynamical attractors from partial observations \cite{takens1980detecting,packard1980geometry}. In the simplest such approach, time-delay embedding, a given univariate measurement time series $x(t)$ is assumed to result from a non-invertible transformation of an underlying multivariate dynamical system $\dot{\v{z}} = \v{f}(\v{z})$ that lies on an attractor. While the dynamical variables necessary to span this attractor are unobserved, time-delay embedding constructs proxy variables using $d_E$ copies of the original measurements some time $\tau$ in the past, resulting in the multivariate time series $\hat{\v{z}} = [x(t), x(t -\tau), ..., x(t - d_E \tau)]$. Theoretical justification for this approach comes from Takens' theorem: if the number of time delays $d_E$ exceeds twice the manifold dimension of the underlying attractor, the resulting time-delay embedding will be diffeomorphic to the original attractor (Box 1) \cite{takens1980detecting}. The surprising aspect of Takens' theorem stems from its apparent contradiction of classical observability: measurements are typically non-invertible and low-rank operators, which discard information in their nullspace \cite{bechhoefer2021control}. Takens' theorem and its variants sidestep this issue by imposing the regularity requirement that the underlying dynamics lie on attractors with well-defined structure---an assumption that, as discussed below, might be recognized as an inductive bias from the perspective of modern statistical learning algorithms. An early success of time delay embeddings was the first experimental detection of a low-dimensional strange attractor as a laboratory flow transitions to turbulence \cite{brandstater1983low}---a critical prediction of the Ruelle-Takens theory of turbulent attractors \cite{ruelle1971nature} (Fig \ref{statespace}A). Delay embeddings subsequently spurred early advances in model-free forecasting \cite{casdagli1989nonlinear,sugihara1990nonlinear,tsonis1992nonlinear} and nonlinear control \cite{ott1990controlling}.

\subsection{Latent representations in time series models}

Classical work on attractor reconstruction bears relevance to present-day statistical forecasting models for time series, which can be seen as generative models $p(\v{x}_t | \v{x}_{t-1}, \v{x}_{t-2}, ...)$ that sample potential future states of a dynamical system conditioned on its past\cite{petropoulos2022forecasting,gershenfeld1999cluster}. Popular \textit{state-space models} for forecasting treat observed data as emissions from an unobserved latent process, such as an underlying attractor or probability distribution \cite{durbin2012time}. In a simplified view, these models decompose time series in three phases: encoding observations into a latent space, propagating the dynamics, and decoding them back into the measurement space. These three stages are made explicit in autoencoders, which parametrize the encoder and decoder using separate artificial neural networks \cite{goodfellow2016deep}. However, even generic statistical learning models for time series, like recurrent neural networks and attention-based transformers, implicitly represent dynamics with hidden variables \cite{girin2021dynamical}. Different statistical time series models may therefore be compared in terms of how they encode and decode time series, how they propagate dynamics in the latent space, and what constraints they apply to each learning stage.

The latent structure found by time series models can reveal dynamical properties not apparent in the original dataset. While attracting inertial manifolds can be shown analytically to exist for particular systems like damped fluid flows \cite{floryan2022data}, reaction-diffusion systems \cite{doering1995applied}, or coupled oscillators \cite{ott2008low}, data-driven detection of latent structures can allow tools from dynamical systems to be applied even in the absence of explicit equations. Recent works have used manifold learning for data-driven nonlinear control, bifurcation detection, and forecasting\cite{blanchard2019learning,cenedese2022data,berry2015nonparametric}. A key theme of recent works involves training autoencoders on high-dimensional dynamical time series, and analyzing the dynamical attractor in the latent space. One recent work decomposes dynamical manifolds through a series of local charts tiling the overlapping latent spaces of several autoencoders \cite{floryan2022data}, an approach reminiscent of classical work on piecewise localized models of chaotic attractors \cite{casdagli1989nonlinear}. Other recent studies analyze local neighborhood incidence to determine the intrinsic dimension of the latent space of autoencoders \cite{gilpin2020deep,chen2022automated}, a concept related to classical neighbor-based methods used to calculate fractal dimensions and select the time-delay embedding dimensions $d_E$ (Box 1) \cite{abarbanel1993analysis}. Consistent with the manifold hypothesis, latent spaces learned by statistical models of high-dimensional dynamical datasets are often contracting; for example, autoencoders trained on videos of weakly turbulent flows can map the dynamics to a low-dimensional latent space associated with exact solutions \cite{page2021revealing}, consistent with inertial manifolds of the underlying partial differential equations. Much like delay embeddings, the dimensionality and model capacity necessary to identify these latent representations often depends on invariants of the underlying dynamics \cite{greydanus2019hamiltonian,linot2020deep}. 

The practical success of contemporary state space models illustrates that seemingly complex dynamics may be generated by low-dimensional latent processes. In this sense, their motivation mirrors earlier efforts to explain seemingly stochastic complex time series in terms of deterministic chaos  \cite{lefebvre1993predictability,sugihara1994nonlinear,tsonis1992nonlinear}. Following the development of time delay embeddings, early approaches to nonlinear forecasting fit the observed data to a dynamical model---either via analytical governing equations or via data-driven methods based on nearest neighbors on the reconstructed attractor \cite{casdagli1992chaos,sugihara1994nonlinear}. \textit{Post hoc} statistical analysis then determined whether including nonlinearity in a fitted model significantly improved the forecast relative to a purely linear model, signalling a deterministic nonlinear component in the dynamics \cite{sugihara1994nonlinear}. Just as classical statistical model selection presumes that the residuals of a fitted model should exhibit uniform scatter, equivalent tests for forecasting models evaluate whether the time series of forecast residuals exhibits no remaining autocorrelation due to unmodelled deterministic dynamics \cite{broock1996test}. Early methods therefore introduced the idea that stochasticity represents intrinsically high-dimensional dynamics driven from unmodelled measurement or process noise, producing degrees of freedom that cannot collapse onto a low-dimensional latent manifold.

When some prior knowledge of a nonlinear system is available, hybrid statistical learning methods directly impose constraints on latent dynamics to ensure consistency with known physical laws. For example, one recent approach encodes high-dimensional time series into a latent space where they obey analytical ordinary differential equations \cite{champion2019data}. These equations can be constrained by restricting the library of possible functions present in the differential equations, or based on known symmetry groups \cite{udrescu2020ai}. Alternative methods such as neural ordinary differential equations do not require the latent differential equation to have an analytic form, only that it can be numerically approximated by an artificial neural network \cite{chen2018neural}. Some works impose constraints through limitations on the architecture of the learning model; for example, Hamiltonian neural networks directly fit differentiable Hamiltonians to data, ensuring that dynamics produced by the learned model heed symplecticity \cite{greydanus2019hamiltonian,choudhary2020physics,toth2019hamiltonian}. When physical constraints are unavailable, other representational constraints can prove informative. In many biological datasets, like recordings of spiking neurons, observed data may be assumed to originate from time-varying stochastic dynamics (like an inhomogenous Poisson process), making the inferred latent dynamics deterministic while the observed dynamics are stochastic \cite{pandarinath2018inferring,koppe2019identifying}.

Modern time series methods therefore navigate a general dichotomy between directly imposing structure (e.g. latent symmetries, symplecticity, distribution families) or inferring these properties from the observed data. The former represents the use of \textit{inductive biases} that shrink the space of possible trained models in order to reduce data-intensivity and errors, at the expense of generalizability. This use of external knowledge about a physical system to tune models along the bias-variance tradeoff echoes classical tradeoffs in nonlinear time series models. Early data assimilation algorithms for chaotic time series, such as nonlinear extensions of the Kalman filter, directly fit the parameters of either known ordinary differential equations or their linearizations \cite{brown1994modeling,julier2004unscented}. The bias-variance tradeoff in these systems appears as rank conditions on the resulting nonlinear fits \cite{reif1999stochastic}. This tradeoff has physical interpretation in early works that used the quality of data-driven models to differentiate low-dimensional chaos from noisy linear dynamics. These works diagnose nonlinearity by comparing the quality of linear and nonlinear models fitted on a given dataset \cite{sugihara1990nonlinear}, with linear dynamics acting as a null-hypothesis with strong inductive bias for stationary time series \cite{kaplan1993model}. These results lead to scaling laws relating the amount of available data, degree of nonlinearity, attractor dimensionality, and the number of time delays required for accurate state-space reconstruction \cite{casdagli1989nonlinear,gershenfeld1992dimension,tsonis1992nonlinear}. Empirical scaling laws relating data volume and model complexity have recently been the focus of intense study by statistical learning practitioners \cite{hoffmann2022training}, suggesting that large generative models of dynamical datasets may eventually obey practical scaling laws backed by theoretical constraints on dynamical systems.

\subsubsection{Lifting linearizes complex dynamics}

The intuition behind time delay embedding---that reparameterization of observed data provides information about unobserved variables---underlies emerging methods at the interface of dynamical systems and machine learning. Originally developed to analyze velocity field measurements in complex fluid flows \cite{schmid2010dynamic}, dynamic mode decomposition seeks to identify the linear transformation mapping a time-delay embedding of a time series onto itself at a later time. For spatially-indexed data like fluid flows, the spectral properties of the resulting linear transformation reveal spatiotemporal motifs such as oscillations, large-scale currents, and other coherent structures \cite{mezic2013analysis,haller2015lagrangian}. Theoretical motivation for this approach stems from recent works showing that nonlinear dynamics become more linear with additional time-delayed variables
\cite{brunton2017chaos,arbabi2017ergodic,kamb2020time}---a concept echoing earlier efforts to unravel nonstationary systems by "overembedding" beyond the prescription of Takens' theorem \cite{hegger2000coping}. These works link dynamic mode decomposition to \textit{Koopman operator theory}, a concept introduced in the early 20th century in the context of ergodic theory for dynamical systems with continuous spectra \cite{koopman1932dynamical}. Given a complex system like a turbulent flow or spiking neuronal array, we can forgo modeling a system in terms of measured variables (velocity fields or individual neuron voltages), and instead “lift” the system to a higher dimensionality than is strictly necessary to fully describe the dynamics. While chaotic systems cannot be linearized with a finite number of lifting transformations, under mild conditions an infinite dimensional transformation exists for which a linear Koopman operator propagates the dynamics \cite{budivsic2012applied}. Recent works have shown empirically that even finite-dimensional approximations of this operator unravel complex dynamical systems, by making them appear quasilinear for extended durations \cite{brunton2017chaos,otto2021koopman}.

However, given a dataset without known governing equations, it is difficult to determine in advance the particular lifting transformations that best approximate the Koopman operator in finite dimensions. Besides time-delays, potential Koopman observables include fixed nonlinear transformations based on known physical symmetries (like spatial Fourier coefficients) \cite{nathan2018applied}; generic nonlinear features like polynomial kernels \cite{williams2015data,nuske2014variational}; custom transformations learned directly from the data using autoencoders or custom kernels \cite{takeishi2017learning,lusch2018deep,wehmeyer2018time}; or transformations identified via equation discovery methods \cite{kaiser2021data}. Because the optimal observables to approximate the Koopman operator are usually unknown {\it a priori}, data-driven methods require regularization and cross-validation to prevent overfitting \cite{bollt2020regularized,li2017extended}. 

Beyond Koopman methods, other emerging operator-theoretic techniques explore the interplay between lifted representations and dynamical complexity. These include data-driven discovery of quadratic forms \cite{qian2020lift}; neural operators for partial differential equations \cite{li2020fourier,karniadakis2021physics}; and works that combine nonlinear transformations of data with symbolic regression of analytical governing equations \cite{chen2022automated,champion2019data}. These frameworks share the theme that dynamical complexity can be unraveled at the expense of increased representational intricacy---echoing tradeoffs between dimensionality and accuracy that underlie Takens' theorem \cite{gershenfeld1992dimension}. Beyond a general tradeoff between cost and accuracy \cite{de2022cost,lu2021learning}, an inefficient choice of lifting transformations undermines interpretability while needlessly increasing computational demands. Similar tradeoffs have been noted in other emerging methods; for example, recently-proposed neural ordinary differential equations use artificial neural networks to construct numerical surrogates for the right hand sides of differential equations \cite{chen2018neural}. Original formulations of these methods struggled to model complex trajectories near kinetic barriers, but later works introduced auxiliary dynamical variables that untangle the trajectories in a lifted space, allowing learned flows to bypass transport obstacles \cite{dupont2019augmented}.  Machine learning practitioners are therefore beginning to confront basic questions regarding transport in dynamical systems with coherent structures inhibiting flow---a return to the original impetus for the development of dynamic mode decomposition, and a demonstration of how dynamical systems theory may inform ongoing practical developments in statistical learning for time series.

\subsection{Outlook for future learning architectures}

The conceptual connections between modern, large-scale time series models, and earlier efforts to discover latent attractors from time delay embeddings, suggest that historical work on chaotic dynamics may continue to provide inductive biases guiding future statistical learning algorithms for time series. While dynamic mode decomposition and Koopman methods have been adapted to broad scientific problems \cite{brunton2022modern,mezic2013analysis,otto2021koopman}, other insights from dynamical systems may prove informative for general time series approaches even beyond the natural sciences. For example, the recently-coined \textit{Hamiltonian manifold hypothesis} notes that because, in principle, all natural videos implicitly illustrate the consequences of physical laws, models trained on sufficiently large datasets will eventually converge to learning latent Hamiltonian dynamics \cite{toth2019hamiltonian}.

On a practical level, constraints drawn from dynamical systems theory have begun to reveal whether the practical success of deep learning arises from unrecognized inductive biases. In deep neural networks, representations of inputs propagate across many layers as they are transformed into output labels or latent representations. Early works formulated multilayer artificial neural networks as continuous-time dynamical systems across layers \cite{pineda1987generalization,chua1988cellular,saad1995line}, a connection that has gained renewed relevance through emerging generative models like neural ordinary differential equations, continuous normalizing flows, and diffusion models \cite{chen2018neural,song2019generative,huguet2022manifold}. Recent works show that dynamics of input representations propagating across layers can even exhibit transient chaos before settling into fixed points associated with output labels \cite{poole2016exponential,schoenholz2016deep}, and that the stretching and folding of input representations across layers gives rise to measures of model expressivity that resemble topological entropy in complex flows \cite{montufar2014number}. Gradient-based training methods for large models implicitly reverse the layerwise dynamics, motivating recent theoretical works describing large models as infinite-dimensional linear dynamical systems \cite{jacot2018neural}, with attendant implications for their ability to learn complex functions \cite{bahri2020statistical}. 

\subsection{Box: Revisiting Embedology: Attractor reconstruction estimates dynamical measures}
% https://community.ams.org/journals/notices/202009/rnoti-p1336.pdf?adat=October%202020&trk=2151&cat=feature&galt=feature

Classic works in nonlinear dynamics describe "embedology," or the process of inferring properties of a dynamical system's attractor, given only low-dimensional time series observations \cite{sauer1991embedology}. To emphasize connections to probabilistic machine learning, we frame embedology as estimating the density of attractor points in phase space. For dissipative chaotic systems in continuous time, this probability distribution often forms a fractal set, which for ergodic systems represents the natural measure $\mu(\v{z})$, or the fraction of time that a long trajectory will spend in the vicinity of a given phase space point $\v{z}$.

Given an observed time series $\{\v{x}_1, \v{x}_2, ..., \v{x}_T\}$, classical attractor reconstruction learns a proxy variable $\v{z}$ associated with the natural measure $p_\theta(\v{z}) \sim \mu(\v{z})$. In classical time delay embedding, the lag operator $\mathcal{L}_\tau[\v{x}_i] \equiv \v{x}_{i - \tau}$ lifts the system to delay coordinates $\v{z}_i \equiv \v{x}_i, \mathcal{L}_\tau[\v{x}_i], \mathcal{L}_\tau^2[\v{x}_i], ..., \mathcal{L}_\tau^{d_E - 1}[\v{x}]$. This reconstruction produces a density estimate $p_\theta(\v{z}) = (T-d_E)^{-1}\sum_{i=1}^{T-d_E} \delta(\v{z} - \v{z}_i)$ with $\theta = \{\tau, d_E\}$, which can be smoothed by centering radial basis functions at each $\v{z}_i$\cite{kantz2004nonlinear,sugihara1994nonlinear}. Motivation for this approach stems from Takens' embedding theorem, a corollary of the Whitney immersion theorem, that states that a time-delay embedding smoothly and invertibly deforms onto the true dynamical attractor as long as $d_E > 2 d_F$, where $d_F$ describes the intrinsic dimensionality of the measure (a non-integer for fractals) \cite{packard1980geometry,takens1980detecting,sauer1991embedology}. However, for most time series $d_F$ and thus $d_E$ are unknown \textit{a priori}; instead $d_E$ and the lag $\tau$ may be treated as learnable parameters $\theta$, with their values determined using heuristic methods. Many methods select the most informative $\tau$ based on local minima of an averaged pairwise similarity measure across the time series $\bar g(\tau) = \langle g(\v{z}_i, \v{z}_{i - \tau}) \rangle_i$; while autocorrelation seems a natural choice, mutual information performs more strongly in practice \cite{abarbanel1993analysis,kantz2004nonlinear}. $d_E$ is typically determined using topological considerations based on neighborhoods in embedding space. Recent works generalize Takens' theorem to multivariate and non-stationary time series \cite{deyle2011generalized,hegger2000coping}, and externally-forced systems with skew-product structure \cite{stark1999delay}.

However, Takens' theorem provides no assurance that time delay embeddings preserve density, $p_\theta(\v{x}) \approx \mu(\v{x})$, a key requirement to accurately sample the system. Many algorithms built upon time-delay embeddings mitigate this issue by performing calculations based on nearest-neighbors, rather than absolute distances in embedding space \cite{sugihara1994nonlinear,kantz2004nonlinear}. Motivated by the Nash embedding theorem, extensions of Takens' theorem have demonstrated conditions under which isometric embeddings can be recovered, often through additional nonlinear transformations or time delays---thus representing a tradeoff between representational dimensionality and accuracy \cite{nash1956imbedding,eftekhari2018stabilizing}. Recovery of the local density has also motivated practical extensions of time delay embeddings based on nonlinear transformations of the lagged coordinates---these include principal components analysis \cite{kamb2020time,brunton2017chaos}, kernel and diffusion map methods \cite{cenedese2022data,berry2015nonparametric}, and artificial neural networks \cite{gilpin2020deep}. 

Evolving a dynamical system produces correlated, not independent, samples from the underlying attractor---suggesting that the measure may be approximated by a set of trajectories, rather than individual points. Unstable periodic orbit theory seeks to group the points comprising the natural measure into exact solutions of the underlying dynamical system, which act as a topological skeleton of the flow \cite{cvitanovic2005chaos}. For dissipative chaotic systems, $\gv\mu(\v{z}) \propto \sum_p \delta(\v{z} - \v{z}_p) \abs{\Lambda_p(\v{z})}^{-1}$, where $p$ indexes a \textit{set of points} $\v{z}_p$ tracing an unstable recurrent solution $\v{z}(t + t_p) = \v{z}(t)$ \cite{grebogi1988unstable,cvitanovic1988invariant,lai1997characterization}. Because chaotic attractors contain no stable points, this sum spans an infinite set of unstable saddles ($t_p \rightarrow 0$) and limit cycles ($t_p > 0$). However, not all solutions influence the dynamics equally, and the stability multiplier $\Lambda_p$ of a given solution denotes its relative instability. Dissipative, hyperbolic chaotic systems exhibit  $\abs{\Lambda_p} > 1$ for all $\v{x}_p$; for saddle points, the stability multipliers may be obtained via linear stability analysis, whereas cycles require averaging across the orbit. Solutions with values closer to one dominate the measure and thus observed dynamics, making them appealing targets for unsupervised learning. Classical methods estimate the dominant unstable periodic orbits directly from dynamical time series by detecting near-recurrences in time-delay embeddings \cite{lathrop1989characterization}. Recently, advances in unsupervised learning and topological data analysis have yielded new methods for detecting unstable periodic orbits in high-dimensional time series \cite{yalniz2021coarse,graham2021exact,bramburger2023data}, prompting new applications of cycle decomposition to complex time series like fluid turbulence \cite{page2021revealing,crowley2022turbulence} and organismal behavior \cite{ahamed2021capturing}.
%check stephens paper: $e^{−\mu_1 p}$ where $\mu_1$ is maximal lyapunov. Sum to weight.
%check hof paper: $\pi_i T$ where $\pi_i$ is the Markov residence time, and $T$ is the period. This is basically a discrete process to jump process. Sum to weight

\begin{figure}
{
\centering
\includegraphics[width=\linewidth]{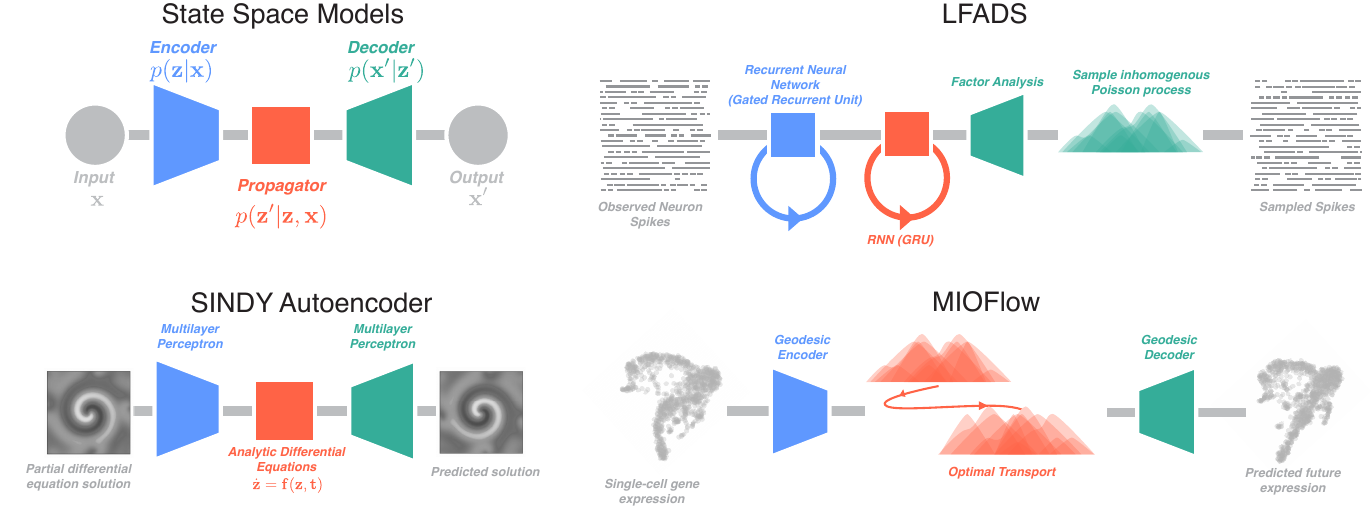}
\caption{
{\bf State space models generate complex dynamics.} (A) Components of a generic state space model. (B) In a variant of Sparse Identification of Nonlinear Dynamics \cite{champion2019data}, multilayer perceptrons deterministically transform high-dimensional observations to a low-dimensional latent space, in which the dynamics are propagated using analytical differential equations learned via sparse regression from a library of known functions. (C) In Latent Factor Analysis via Dynamical Systems \cite{pandarinath2018inferring}, neuron spiking time series are deterministically encoded into latent initial conditions, which are evolved using a second recurrent neural network, and then decoded into latent factor time series. These latent factors parameterize the stochastic firing rate of an inhomogenous Poisson process.  (D) In Manifold Interpolating Optimal-Transport Flows \cite{huguet2022manifold}, high-dimensional gene expression measurements are encoded to a latent distribution that preserves the manifold diffusion distance. The latent measure is then propagated with optimal transport.
}
\label{state}
}
\end{figure}

\section{Compressibility and minimal dynamical generators}
%+ Lead with historical context
%+ Then ML connections
%+ Then work at the interface
%+ Then outlook

Latent space representations imply that the apparent complexity of a dynamical process may depend on the choice of measurement coordinates. Yet Pesin's formula associates entropy production with chaotic dynamics, implying that some aspects of chaos are irreducible: intuitively, no invertible reparametrization can map a strange attractor to a limit cycle. Moreover, entropy production often has observable effects, like heat production, that are independent of representation \cite{conte2019thermodynamic,landauer1961irreversibility}. Classical works explore the irreducibility of chaos in the context of symbolic dynamics, which consider the computational properties of continuous systems under discrete coarse-graining \cite{shaw1981strange,morse1938symbolic}. A continuous-valued dynamical time series may be converted into a symbolic series by partitioning phase space, and then analyzing the properties of the symbol sequence produced by recording the partition label whenever the deterministic dynamics cross a boundary. Pesin's formula implies that this sequence is non-repeating for any nontrivial partitioning of a continuous-time chaotic system. However, because no deterministic finite-state automaton can exhibit non-recurrent dynamics, only a stochastic automaton can describe the symbol sequence produced by partitioning a deterministic chaotic system \cite{moore1990unpredictability,crutchfield1989inferring}. Symbolic dynamics thus links the properties of analogue dynamical systems to digital computers, exchanging determinism for brevity.

An early motivation for symbolic dynamics was identifying computational equivalence among systems\cite{bowen1973symbolic,metropolis1973finite}: setting aside differences in representation, are certain dynamical systems functionally identical? For example, the celebrated period-doubling cascade provides a universal description of bifurcations leading to chaos across diverse systems, ranging from turbulent flows to ecological population fluctuations \cite{crutchfield1982symbolic,hao1991symbolic}. The dynamics preceding a given period doubling bifurcation can be mapped to those following the bifurcation via a renormalization operation,\cite{feigenbaum1979universal} the repeated application of which drives the system towards an accumulation point where the period diverges and chaos emerges. Symbolic dynamics provides an alternative view of this process, in which periodic dynamics correspond to a two-state automaton implementing a discrete shift on a symbolic register \cite{hao1991symbolic}. Each bifurcation doubles the number of unique states (and thus memory requirements) of the automaton, but renormalization decimates the register and renders the automaton invariant. At the accumulation point, the periodicity and thus memory requirements diverge, and the asymptotic entropy production becomes nonzero. Similar analyses based on symbolization were used by early practitioners to identify universal structure in diverse systems such as kicked rotors and chaotic scattering \cite{lewis1992nonlinear,hao1989elementary}, as well as to map experimental datasets onto characteristic minimal systems \cite{daw2003review}. Other contemporaneous works even sought to enumerate and categorize minimal symbolic dynamical systems based on their ability to support computation \cite{langton1990computation,wolfram1984universality}.

\subsection{Models that learn to distill dynamics}

\begin{figure}
{
\centering
\includegraphics[width=\linewidth]{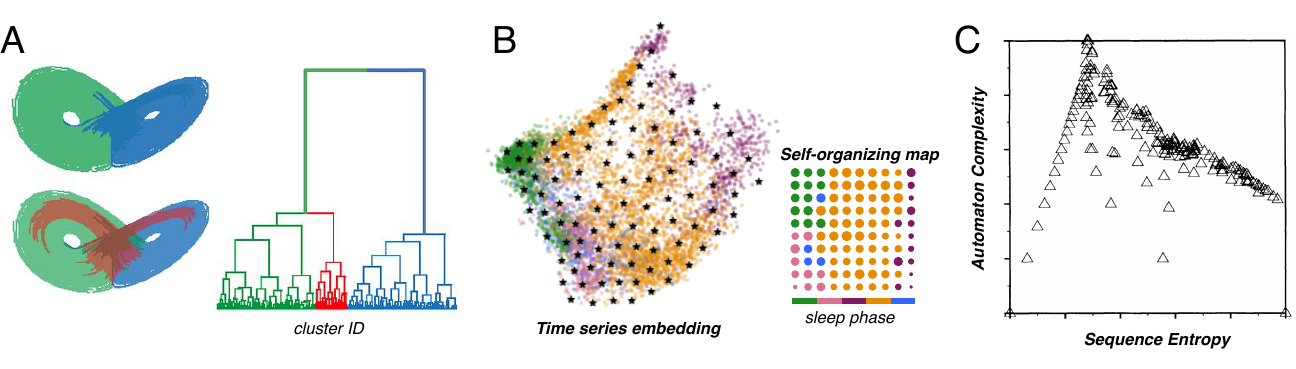}
\caption{
{\bf Latent discretization and interpretability.}  (A) Successive stages of an adaptive approximation algorithm that fits locally-linear dynamics to parts of the phase space of a chaotic system. (B) A continuous-valued learning model that creates a discrete, latent self-organizing map from a continuous time series of sleep recordings. (C) The topological complexity of probabilistic automata fitted to a dynamical map across a range of chaotic and periodic regimes, plotted against the entropy of the time series. The most structurally-complex automaton occurs when the dynamics exhibit intermediate entropy.
\textit{Panel A is reprinted from Costa, Ahamed, Stephens 2019. Panel B is reprinted from Huijben et al. 2023. Panel C is reprinted from Young \& Crutchfield, 1989.\cite{costa2019adaptive,huijben2023som,crutchfield1989inferring}.}
}
\label{discrete}
}
\end{figure}

% Imposing discreteness for interpretability reasons
The motivation behind symbolic dynamics---reducing systems down to their essential components---seemingly contradicts current trends of scaling general-purpose learning models to ever-larger datasets. However, recent works on model compression and distillation revisit the original ambitions of symbolic dynamics. When fitting a many-parameter learning model to a given dataset, patterns within the original data (such as symmetries or stereotypy) may be revealed through analysis of the trained model. For example, many state space time series models directly map continuous-time observations to discrete modes of the underlying system (Fig. \ref{state}). In particular, hidden Markov models treat continuous observations as emissions from probability distributions conditioned on sequences of discrete internal states \cite{durbin2012time}. Likewise, switching linear models fit piecewise linear operators to subsets of a system's full phase space, thereby approximating the global dynamics through a series of switches among local linear maps (Fig. \ref{discrete}A) \cite{casdagli1989nonlinear,ghahramani2000variational,fox2008nonparametric,smith2021reverse}. These approaches have proven particularly successful for datasets like organismal behavior and speech patterns, where high measurement dimensionality meets low dynamical dimensionality due to biomechanical constraints \cite{johnson2016composing,costa2019adaptive}. In such cases, latent discretization provides interpretability \cite{johnson2016composing,krakovna2016increasing}; for example, in continuous-time recordings of organismal behavior or neuronal activity, the latent variable sequence indicates distinct cognitive imperatives \cite{smith2021reverse,mudrik2022decomposed}. 

Natural images and other real-world datasets often span low-dimensional manifolds relative to their feature set \cite{fefferman2016testing}. As a result, many large-scale generative learning models are designed to map between complex datasets and simplified latent representations; for example, variational autoencoders use artificial neural networks to map training data to a tractable probability distribution, which can then be sampled to generate new surrogate data. This latent space can therefore illuminate the inner workings of the learning model, even when the learned transformation itself between ambient and latent spaces remains opaque. For example, several recently-developed architectures apply constraints during learning that cause variational autoencoders to learn a quantized latent space \cite{van2017neural,devaraj2020symbols,rasul2022vq,falck2021multi}, in which patterns in the input data correspond to discrete entries within a latent codebook. These discrete modes reveal clusters of related training examples, making these models a generalization of classical self-organizing maps (Fig. \ref{discrete}B) \cite{fortuin2018som,kohonen1990self}. Beyond interpretability, imposing discretization helps large generative models avoid \textit{posterior collapse}, a limitation of autoregressive generation in which the model begins ignoring the latent state space, and instead relies only on the decoder to determine its output---thereby reducing the complexity of the generated samples \cite{van2017neural,blei2017variational}. To identify this and other failure modes, recent works propose using entropy production to identify miscalibration in generative models \cite{braverman2020calibration}.

A limitation of modern overparameterized learning models stems from degeneracy: many different trained models may exhibit equivalent performance on a given task, thereby precluding systematic comparison of models across tasks or model architectures. However, simplified latent representations of learned dynamics can reveal commonalities across learning models, echoing the computational primitives sought by symbolic dynamics. Quantized latent states can reveal the internal logic of black-box neural networks, by allowing their internal grammar to be probed with post-hoc analysis
\cite{smith2021reverse,falck2021multi,tschannen2018recent,jang2017categorical}. For example, traditional recurrent neural networks are provably capable of encoding arbitrary continuous-time dynamical attractors \cite{funahashi1993approximation} and even discrete logic \cite{neto1997turing,kaiser2015neural}, given sufficient training data and model scale. Trained black-box recurrent learning models can be analyzed by fitting probabilistic automata to their latent dynamics \cite{weiss2019learning}, and recent works show that newer classes of generative learning models, like attention-based transformers, may outperform earlier architectures because they can internally represent more sophisticated grammars \cite{michalenkorepresenting,resnick2019capacity,liu2022transformers}. Rather than analyzing continuous-valued learning models post-training, some methods instead impose discrete dynamics directly through architectural constraints on the model. Emerging neuro-symbolic approaches combine the trainability of continuous-valued learning models with the representational guarantees of exact symbolic procedures, like digital logic or arithmetic, by incorporating the latter within separate modules \cite{tsamoura2021neural,daniele2022deep,trask2018neural,kaiser2015neural,yik2023neurobench}. 

A drawback to combining discrete operations with continuous-valued model parameters stems from the difficulty of computing gradients of discrete states, which complicates training of large models with gradient descent. This limitation mirrors a common shortcoming of symbolic dynamical systems, for which non-differentiability precludes the application of mathematical tools such as linear stability analysis---leading some early practitioners to view symbolic dynamics as intrinsically computational objects, that can only be understood through direct simulation \cite{neumann1966theory,wolfram1983statistical,langton1990computation}. End-to-end trainable learning models containing discrete modules frequently use straight-through estimators, in which symbolic components are treated as identity functions when computing gradients of the model error with respect to its parameters \cite{van2017neural}. Other recent works bypass gradient-based training entirely, instead directly modifying the latent code within continuous-valued neural networks, in order to program the dynamics to perform discrete computations \cite{gilpin2019cellular,kim2023neural}. 

Taken together, these works illustrate how implicit or imposed symbolization within trainable learning models allows the generation of more complex and interpretable dynamics. Future works may use symbolic methods to automatically identify universality in generators of time series found by different trained learning models. For example, systems biology often requires comparison of gene regulatory dynamics across multiple organisms. While these measurements vary widely based on differences in imaging modalities and fluorescent reporters, symbolic distillation could identify shared latent dynamical motifs arising from orthologous regulatory structures \cite{wong2020gene}. A key concept that may inform future developments in generative modelling is \textit{unifilarity} \cite{crutchfield1989inferring}. While a given time series can be mapped to multiple possible state space models, a unifilar representation comprises the minimal maximally predictive generator for the dynamics---thus facilitating comparison of generators across systems \cite{crutchfield2012between,ephraim2002hidden,marzen2017nearly}. Recent improvements in inference methods thus represent a key step towards extracting minimal latent generators that can be compared across datasets and trained learning models \cite{strelioff2014bayesian,marzen2017structure,pfau2010probabilistic}.

\subsubsection{Measuring entropy production from data}

Classical work on symbolic dynamics bears relevance to emerging interdisciplinary problems for which entropy production has physical implications. Recent studies have sought to identify macroscopic signatures of microscale nonequilibrium processes directly from experimental data \cite{battle2016broken,lucente2022inference,frishman2020learning,skinner2021improved}. Equilibrium thermodynamic systems exhibit detailed balance, in which the net flux between any pair of microstates equals zero. In contrast, active systems like biological structures dissipate energy at the microscale, thus producing apparent violations of detailed balance at larger scales. Recent works have shown that finite-resolution time series measurements of such systems, such as those produced by video microscopy, exhibit signatures of these microscale effects when quantized in the spatial or frequency domains \cite{battle2016broken}. Nonequilibrium behaviors manifest as net circulation in the phase space of the coarse-grained data---in contrast to detailed balance, in which probability currents vanish. These methods have successfully identified mesoscopic nonequilibrium states in diverse systems ranging from the locomotory states of swimming cells \cite{wan2018time,larson2022unicellular} to oxygen levels in the brain during cognitive exertion \cite{lynn2021broken}. 

Because executing computations in finite time and noisy environments requires energy dissipation, the physical entropy produced by nonequilibrium thermodynamic flows can be related to the information-theoretic entropy production associated with symbolic dynamics. Recent works exploit this relationship by symbolizing active systems using binary occupancy statistics of individual degrees of freedom, and then applying existing digital compression algorithms to efficiently measure the difference in probability between forward and backward sequences, a measure of irreversibility \cite{martiniani2020correlation,ro2022model,nardini2017entropy}. Nonequilibrium steady states can thus reveal information about minimal dynamical systems underlying biological motifs. Emerging works on data-driven detection of nonequilibrium flows have begun to infer computational primitives of the underlying living systems \cite{tkavcik2016information,frishman2020learning}, particularly in systems known to execute computations such as neuronal ensembles \cite{lynn2021broken,lynn2022decomposing} or cellular decision-making \cite{larson2022unicellular,bauer2021trading,mattingly2021escherichia,wong2020gene}. 

% Thermodynamic computing and irreversible computation
These works provide biophysical motivation to revisit classical questions regarding the physical nature of information in dynamical systems that support computation \cite{moore1990unpredictability,feynman2018feynman}. In computers operating at finite temperatures, small-scale thermal fluctuations represent a precision floor. If computation is implemented with chaotic dynamics, errors cascade from small to large scales at a rate that depends only on the Lyapunov exponents of the deterministic dynamics---and not on the temperature itself \cite{shaw1981strange}. The mutual information between the initial and final states of a chaotic system---deemed the \textit{transinformation} in early works---therefore decays over time, a form of information erasure that underlies the effective irreversibility of chaos \cite{pompe1986state}. Work predating the widespread study of chaos established a minimal thermodynamic cost for information erasure \cite{landauer1961irreversibility}, a connection that influenced later efforts to understand computation in chaotic systems \cite{landauer1987computation}. Recent works revisit these concepts by formalizing the nonequilibrium thermodynamics of information processing systems \cite{still2012thermodynamics,conte2019thermodynamic,tkavcik2016information}.

Nonequilibrium thermodynamics has influenced the recent development of practical generative models for complex datasets \cite{adhikari2023machine,li2023measuring}. A leading contemporary approach to natural image generation is diffusion models, which learn to iteratively invert a diffusive flow connecting a tractable latent distribution to the observed distribution of natural images \cite{song2019generative}. Training a diffusion model on natural images consists of gradually adding high-dimensional noise pixelwise to each input image, while simultaneously training a set of denoising learning models to invert each incremental noise addition. After training, synthetic images may be generated by sampling an image of random noise, and then applying the sequence of trained denoising models \cite{ho2020denoising}. Early works on diffusion models noted that this sequence of intermediate operations comprises a nonequilibrium flow \cite{sohl2015deep}, and that generation of new images requires weighted nonequilibrium sampling of rarer trajectories that lead to realistic natural images. Recent empirical studies assessing the quality of large, autoregressive generative models observe long-time decay in mutual information between input states and generated outputs \cite{braverman2020calibration}, a phenomenon comparable to classical transinformation erasure in chaotic systems \cite{shaw1981strange}. By analogy to the data processing inequality, transinformation decay implies that, given knowledge of a dynamical system at finite precision, no statistical learning model can recover predictive information about the initial conditions, once a sufficient number of Lyapunov times have elapsed \cite{tkavcik2016information,pompe1986state}.

%Statistical learning models heed the well-known data processing inequality: a model cannot increase the information content in a training dataset, it can only identify structure most relevant to given task \cite{tkavcik2016information}. Transinformation decay represents an equivalent constraint for dynamical systems: given knowledge of a dynamical system at finite precision, no statistical learning model can recover predictive information about the initial conditions, once sufficiently-many Lyapunov times have elapsed \cite{pompe1986state}.

%(B) A vector-quantized variational autoencoder (VQ-VAE) maps high-dimensional continuous-valued cell images to discrete phases of the cell cycle.

\subsection{Box: Generative statistical models of complex dynamics}

A generative model $p_\theta(\mathbf{x}_{1:T})$ for a time series $\v{x}_{1:T} \equiv \v{x}_1, \v{x}_2, ..., \v{x}_T$ has the form \cite{girin2021dynamical,blei2017variational,foti2014stochastic},
\begin{equation}
%p_\theta(\mathbf{x}_{1:T}) = \int p_\theta(\v{x}_{1:T}, \v{z}_{1:T})  \, d\mathbf{z}_{1:T}
p_\theta(\mathbf{x}_{1:T}) = \int p_\theta(\v{x}_{1:T} | \v{z}_{1:T}) p_\theta(\v{z}_{1:T}) \, d\mathbf{z}_{1:T}
\label{generator}
\end{equation}
where $p(\mathbf{z}_{1:T})$ denotes the prior distribution of the model's latent state sequence, which here matches the temporal resolution of the measured time series; $p_\theta(\v{x}_{1:T} | \v{z}_{1:T})$ denotes the likelihood of an observed time series given a latent sequence; and $\theta$ denotes the trainable parameters. Many time series exhibit a characteristic timescale $\tau < T$ over which future values become decorrelated from past; for deterministic chaotic time series this approximately comprises the Lyapunov time $\tau \approx \lambda_\text{max}^{-1}$. In this case, the likelihood exhibits \textit{conditional independence}, in which all relevant dependence between past and future values of $\v{x}$ are captured in the $\tau$ most recent timepoints, 
\[
	p_\theta(\v{x}_{1:T} | \v{z}_{1:T}) = \prod_{t=1}^{\tau} p_\theta(\v{x}_t | \v{z}_{1:t}) \prod_{t=\tau+1}^{T} p_\theta(\v{x}_t | \v{z}_{t-\tau+1:t})
\]
%\textcolor{red}{How to incorporate dependence on past x values?}
Moreover, the latent prior becomes \textit{autoregressive},
\[
	p_\theta(\v{z}_{1:T}) = p_\theta(\v{z}_{1:\tau}) \prod_{t=\tau+1}^{T} p_\theta(\v{z}_t | \v{z}_{t-\tau:t-1})
\]
Inserting this into \eqref{generator} and rearranging terms yields a general expression for a generative time series model,
\begin{equation}
p_\theta(\mathbf{x}_{1:T}) = \int 
\left[
p_\theta(\mathbf{z}_{1:\tau}) \prod_{t=1}^{\tau} p_\theta(\mathbf{x}_t | \mathbf{z}_{1:t}) 
 \right]
\left( 
\prod_{t=\tau+1}^{T} p_\theta(\mathbf{x}_t | \mathbf{z}_{t-\tau+1:t})  p_\theta(\mathbf{z}_t | \mathbf{z}_{t-\tau:t-1}) 
\right) \, d\mathbf{z}_{1:T}
\label{emissiontransmission}
\end{equation}
where the first bracketed term represents the initial conditions at the start of the time series. The second term comprises two parts: an \textit{emission} model for $\v{x}$ based on the past $\tau$ values of $\v{z}$ and $\v{x}$, and a \textit{transition} model implementing the latent dynamics. In most practical applications, $\tau$ is unknown \textit{a priori} and instead represents an adjustable hyperparameter of the model---in forecasting, $\tau$ is the \textit{lookback window}, while in other machine learning applications $\tau$ is the \textit{context length} akin to a working memory. In applications like data smoothing and assimilation, the latent series $\v{z}_{1:T}$ may represent the variable of practical interest, while forecasting seeks to directly sample future states of $\v{x}$. For a traditional hidden Markov model, the transmission kernel simplifies to $p_\theta(\v{z}_t | \v{x}_{t-\tau:t-1} , \v{z}_{t-\tau:t-1}) = p_\theta(\v{z}_{t} | \v{z}_{t-1})$. For models implementing deterministic latent dynamics (for example, latent ordinary differential equations), the transmission becomes $p_\theta(\mathbf{z}_t | \mathbf{x}_{t-1}, \mathbf{z}_{t-1}) = \delta(\mathbf{z}_t - \v{F}(\mathbf{z}_{t-1}, \mathbf{x}_{t-1}))$, where the flow map $\v{F}$ propagates $\v{z}_{t-1}$ to $\v{z}_{t}$, potentially under external forcing by $\v{x}$.

Training probabilistic models requires maximizing the marginal log-likelihood $\log p_\theta(\mathbf{x}_{1:T})$ of the training data under the learned parameters $\theta$. If the underlying dynamical system is linear and all conditional probabilities and process noise are Gaussian, then \eqref{emissiontransmission} reduces to a form similar to \eqref{entgauss}, and the resulting model represents the Kalman filter \cite{kalman1960new,roweis1999unifying}. The likelihood of other classical state space models like hidden Markov models may be trained iteratively with expectation-maximization procedures \cite{durbin2012time,blei2017variational}. However, for many complex time series, $p_\theta(\v{z}_{1:T} | \v{x}_{1:T})$ becomes difficult to sample, and so many recent works approximate the posterior distribution using artificial neural networks. Supervised training of forecasting models, like recurrent neural networks or attention-based transformers, requires comparing the model's generated predictions against ground truth future values---minimizing the forecast error therefore maximizes the training data likelihood. In unsupervised settings, training can instead proceed by comparing a given training example to one sampled from a nearby latent space location. In generative adversarial networks, training feedback is derived by passing the generator outputs to a separate discriminator that attempts to distinguish true versus sampled points \cite{goodfellow2014generative}. 

The marginal likelihood $p_\theta(\mathbf{x}_{1:T})$  is typically difficult to compute, and so recent approaches like variational autoencoders instead optimize a lower bound on the marginal log-likelihood called the evidence lower bound \cite{kingma2014semi}. These approaches introduce a second trainable model that approximates the posterior $q_\phi(\v{z}_{1:T} | \v{x}_{1:T}) \approx p_\theta(\v{z}_{1:T} | \v{x}_{1:T})$. These models are then trained to minimize a variational bound on the negative log-likelihood,
\[
	-\log p_\theta(\v{x}_{1:T}) 
	= - \log \int p_\theta(\v{x}_{1:T} , \v{z}_{1:T}) \,d\v{z}_{1:T}
	\leq \mathbb{E}_q\left[		-\log\dfrac{p_\theta(\v{x}_{1:T}, \v{z}_{1:T})}{q_\phi(\v{z}_{1:T} | \v{x}_{1:T})} \right]
\]
where the latter term arises from Jensen's inequality. We equate this expression with a loss function and rearrange terms to reveal an entropy-like expression,
\[
	\mathcal{L}_{\theta, \phi}(\v{x}_{1:T}) = -\mathbb{E}_q\left[\log p_\theta(	 \v{x}_{1:T}, \v{z}_{1:T}) \right] + \int q_\phi(\v{z}_{1:T} | \v{x}_{1:T}) \log q_\phi(\v{z}_{1:T} | \v{x}_{1:T}) \, d\v{z}_{1:T}
\]
We next apply the same conditional independence assumptions used to derive \eqref{emissiontransmission}. We neglect the boundary term by assuming that the loss depends negligibly on the first $\tau \ll T$ timepoints,
%\[
%	\sum_{t =1}^T \mathbb{E}_{\v{z}\sim q} 
%	\left[		
%		\log p(\v{x}_t | \v{z}_t) 
%		+ \log p(\v{z}_t | \v{z}_{1:t-1})
%		- \log q_\phi (\v{z}_t | \v{z}_{1:t-1}, \v{x}_{1:t})
%	\right]
%\]
\begin{equation}	
	\mathcal{L}_{\theta, \phi}(\v{x}_{1:T}) \approx
	-\sum_{t=\tau+1}^T \mathbb{E}_q 
	\left[		
		\log p_\theta(\mathbf{x}_t | \mathbf{z}_{t-\tau+1:t})
%		+ \log p_\theta(\mathbf{z}_t | \mathbf{z}_{t-\tau:t-1}) 
%		- \log q_\phi (\v{z}_t | \v{z}_{t-\tau:t-1}, \v{x}_{t-\tau+1:t})
- \log\left(\dfrac
{q_\phi (\v{z}_t | \v{z}_{t-\tau:t-1}, \v{x}_{t-\tau+1:t})}
{p_\theta(\mathbf{z}_t | \mathbf{z}_{t-\tau:t-1})}
\right)
	\right].
\label{splitloss}
\end{equation}
The loss therefore splits into a series of separate contributions from each $\tau$-timepoint window, resembling classical state space factorization of chaotic time series.\cite{abarbanel1993analysis} During training, the emission term $p_\theta(\v{x} | \v{z})$ and approximate transition term $q_\phi(\v{z}_t | \v{z}_{t-1}...)$ may be parameterized with models such as attention-based transformers or recurrent neural networks \cite{tang2021probabilistic}. After training, forecasts may be generated autoregressively using \eqref{emissiontransmission}. The first term in \eqref{splitloss} corresponds to a maximum likelihood term that encourages accurate reconstruction of the training data, while the second term minimizes the Kullback-Leibler divergence between the true latent transition term $p_\theta$ and its trainable surrogate $q_\phi$. A similar information-theoretic expression appears in classical measures of synchronization in chaotic systems \cite{schreiber2000measuring,bollt2003review,baptista2005chaotic}, and recent works propose generalized synchronization between true and latent dynamics as a potential learning mechanism for recurrent neural networks \cite{lu2020invertible}. This effect may explain recent empirical works demonstrating that modern recurrent neural networks can successfully forecast chaotic systems $\sim\!\!10$ Lyapunov times into the future \cite{pathak2018model,yik2023neurobench,gilpin2021chaos}.

\section{Outlook: Complexity versus entropy}

Much as early computers and the resulting visualizations of fractals inspired excitement in applying chaos to other fields \cite{campbell1985experimental}, recent advances in statistical learning have sparked renewed interest in classical ideas from nonlinear dynamics. Connections between these fields range from physics-based inductive biases on latent representations in generative learning models, to the identification of minimal dynamical generators underlying complex time series.

Future works may implicate fundamental relationships between the observability and representability of complex dynamics. Early efforts to relate chaos to computational principles related the apparent entropy of a system to the complexity of its underlying representation \cite{crutchfield1989inferring,feynman2018feynman}. A system settling to a fixed point or limit cycle eventually ceases to produce new information because its attractor has been fully observed after a finite observation period \cite{crutchfield2012between,feldman2008organization}. Conversely, a completely stochastic system like a random number generator seemingly produces information, but without any underlying structure. The complexity of a system's generator plotted against the entropy of its outputs therefore exhibits non-monotonicity with an intermediate peak---suggestively termed the "edge of chaos" by some practitioners---that represents systems that can, at different times, switch between fully-ordered and seemingly random outputs (Fig. \ref{discrete}C). Early works considered whether this edge represents those systems capable of supporting information processing and intelligence \cite{feldman2008organization,langton1990computation,mitchell1994dynamics}, a concept revisited in recent studies analyzing the capacity of modern statistical learning models \cite{carroll2020reservoir,fajardo2022fundamental,krishnamurthy2022theory,mikhaeil2022difficulty,marzen2023complexity}.

As the scale and quality of generative learning models improves, structural complexity versus data randomness may emerge as an observable relationship between problem difficulty and model selection.  A complexity-entropy relation could describe the intricacy of latent representations learned by large models in unsupervised settings, or the complexity of the underlying architectures necessary to achieve a given accuracy on supervised learning problems. This dynamical refinement of the bias-variance tradeoff could inform future developments, bridging Wheeler's physical bits with the practicalities of modern large-scale learning systems.

\section*{Acknowledgments}

We thank Henry Abarbanel and Harry Swinney for informative discussions. This project has been made possible in part by grant number DAF2023-329596 from the Chan Zuckerberg Initiative DAF, an advised fund of Silicon Valley Community Foundation.

\clearpage

% 100 - 200 references are okay
\bibliography{chaos_cites} 
\bibliographystyle{naturemag}

\end{document}